\documentclass[letterpaper, 10 pt, conference]{ieeeconf}  % Comment this line out if you need a4paper

\IEEEoverridecommandlockouts                              % This command is only needed if 
\usepackage{times} % assumes new font selection scheme installed
\usepackage{amsmath} % assumes amsmath package installed
\usepackage{amssymb}  % assumes amsmath package installed

\usepackage{amsmath,amsfonts,bm}

\def\eqref#1{equation~\ref{#1}}
\def\1{\bm{1}}

\def\vf{{\bm{f}}}

\def\vp{{\bm{p}}}
\def\vq{{\bm{q}}}

\def\vt{{\bm{t}}}

\def\mM{{\bm{M}}}

\DeclareMathAlphabet{\mathsfit}{\encodingdefault}{\sfdefault}{m}{sl}
\SetMathAlphabet{\mathsfit}{bold}{\encodingdefault}{\sfdefault}{bx}{n}

\usepackage[hidelinks]{hyperref}
\usepackage{cleveref}
\usepackage{url}

\usepackage{amsfonts}
\usepackage{xcolor}
\usepackage{comment}

\usepackage{graphicx}
\usepackage{subfig}
\usepackage{adjustbox}
\usepackage{tabularx}
\usepackage{wrapfig}
\usepackage{booktabs}
\usepackage{cite}
\usepackage{caption}
\usepackage{dblfloatfix}

\title{\LARGE \bf
PokeFlex: A Real-World Dataset of \\ Volumetric Deformable Objects for Robotics
}

\author{Jan Obrist$^{*,1}$, Miguel Zamora$^{*,1}$, Hehui Zheng $^{*,2,3}$, Ronan Hinchet$^{2}$, \\ Firat Ozdemir$^{5}$,  Juan Zarate$^{4}$, Robert K. Katzschmann$^{2,3}$ and Stelian Coros$^{1}$% <-this % stops a space
\thanks{$^{1}$Computational Robotics Lab, ETH Zurich}%
\thanks{$^{2}$Soft Robotics Lab, ETH Zurich}%
\thanks{$^{3}$ETH AI Center, ETH Zurich}%
\thanks{$^{4}$Advanced Interactive Technologies Lab, ETH Zurich}%
\thanks{$^{5}$Swiss Data Science Center, ETH Zurich \& EPFL}% 
\thanks{$^{*}$Equal contribution. }% 
\thanks{This work is supported by the SDSC Grant entitled 'C22-08: Data-Driven Inference of Mesh-based Representations for Deformable Objects from Unstructured Point Clouds'}
}

\ifdefined\showColumnMargins 
    \usepackage{showframe}
    \usepackage{etoolbox}
    
    \newlength\Fcolumnseprule
    \makeatletter
    \newcommand\ShowInterColumnFrame{
    \patchcmd{\@outputdblcol}
      {{\normalcolor\vrule \@width\columnseprule}}
      {\vrule \@width\Fcolumnseprule\hfil
        {\normalcolor\vrule \@width\columnseprule}
        \hfil\vrule \@width\Fcolumnseprule
      }
      {}
      {}
    }
    
    \makeatother
    \ShowInterColumnFrame
\fi

\usepackage{fancyhdr}
\begin{document}
\fancyhead{} % clear all header fields
\fancyhead[C]{\textbf{The performance of new graduates}}

\newcommand{\Comment}[3]
{
\textcolor{#3}{#1: #2}
}

\newcommand{\JO}[1]
{
\Comment{JO}{#1}{red}
}

\newcommand{\MZ}[1]
{
\Comment{MZ}{#1}{cyan}
}

\newcommand{\JZ}[1]
{
\Comment{JZ}{#1}{teal}
}

\newcommand{\HZ}[1]
{
\Comment{HZ}{#1}{blue}
}

\newcommand{\FO}[1]{\Comment{FO}{#1}{violet}}

\newcommand{\reb}[1]{{\color{black}{#1}}}

\makeatletter
\let\@oldmaketitle\@maketitle% Store \@maketitle
\renewcommand{\@maketitle}{
  Preprint 
  \vspace{-5mm}
  \@oldmaketitle% Update \@maketitle to insert...
  \begin{center}
  \vspace{10pt}
    \includegraphics[width=0.85\linewidth]{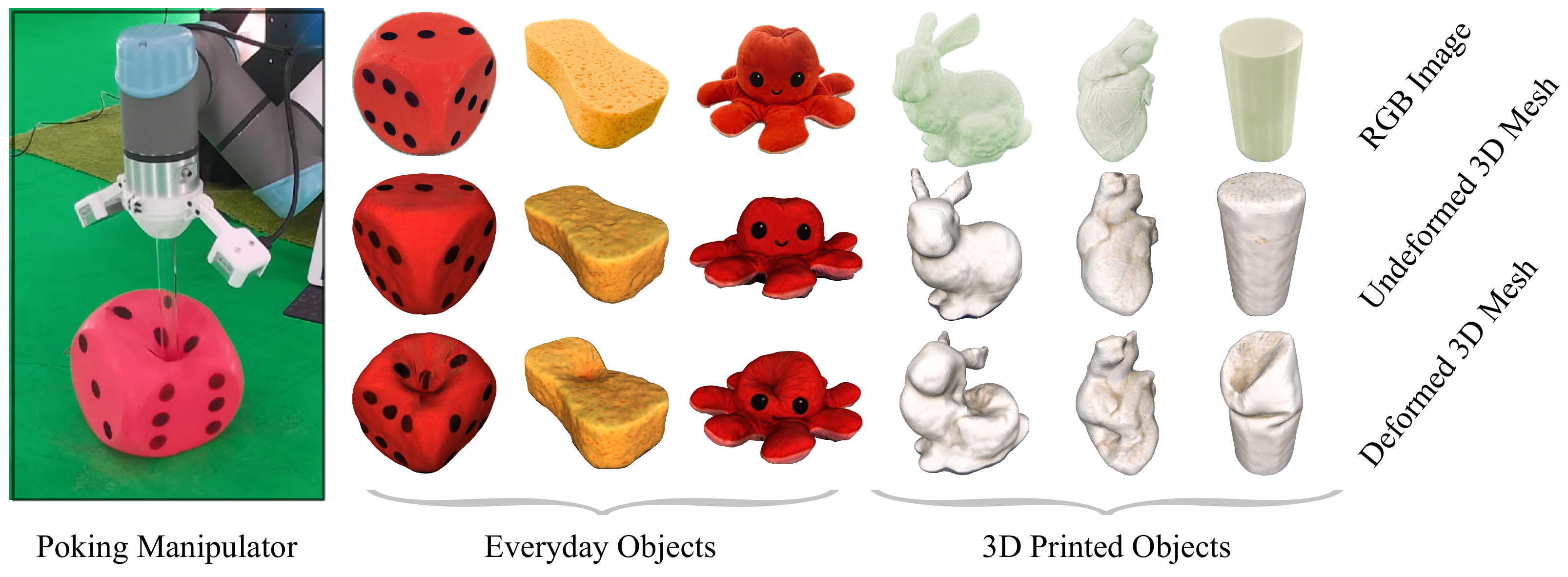}
    % \vspace{0.3cm}
    \captionsetup{font=footnotesize}
    \captionof{figure}{PokeFlex captures the deformability of various everyday and 3D-printed objects, as illustrated by the poking manipulator on the \textbf{Left}.  On the \textbf{Right}, the \textbf{Top Row} contains segmented RGB images of selected objects. The \textbf{Middle Row} shows reconstructed objects in an undeformed state. The \textbf{Bottom Row} provides reconstructed 3D-textured meshes of deformed objects.  For an overview of all the objects, see \Cref{fig:objects}. }
    \label{fig:overview}
    \vspace{-5mm}
  \end{center}\bigskip
}
\makeatother
\maketitle
\addtocounter{figure}{-1}
\thispagestyle{empty}
\pagestyle{empty}

% \maketitle
% \thispagestyle{empty}
% \pagestyle{empty}

%%%%%%%%%%%%%%%%%%%%%%%%%%%%%%%%%%%%%%%%%%%%%%%%%%%%%%%%%%%%%%%%%%%%%%%%%%%%%%%%
% \begin{figure}[h!]
% \centering
% %\vspace{-2.5\baselineskip}
% \includegraphics[width=0.95\linewidth]{images/fig1_v5.pdf}
% %\vspace{-0.5\baselineskip}
% \caption{PokeFlex captures the deformability of various everyday and 3D-printed objects, as illustrated by the poking manipulator on the \textbf{Left}.  On the \textbf{Right}, the \textbf{Top Row} contains segmented RGB images of selected objects. The \textbf{Middle Row} shows reconstructed objects in an undeformed state. The \textbf{Bottom Row} provides reconstructed 3D-textured meshes of deformed objects.}
% \label{fig:overview}
% \end{figure}

\begin{abstract}

% Motivation
Data-driven methods have shown great potential in solving challenging manipulation tasks; however, their application in the domain of deformable objects has been constrained, in part, by the lack of data. 
% Goal
To address this lack, we propose PokeFlex, a dataset featuring real-world multimodal data that is paired and annotated. The modalities include 3D textured meshes, point clouds, RGB images, and depth maps. Such data can be leveraged for several downstream tasks, such as online 3D mesh reconstruction, and it can potentially enable underexplored applications such as the real-world deployment of traditional control methods based on mesh simulations.
% Challenge
To deal with the challenges posed by real-world 3D mesh reconstruction, 
% Approach
we leverage a professional volumetric capture system that allows complete 360\textdegree \,reconstruction. PokeFlex consists of 18 deformable objects with varying stiffness and shapes. Deformations are generated by dropping objects onto a flat surface or by poking the objects with a robot arm. Interaction \reb{wrenches and contact locations} are also reported for the latter case. 
Using different data modalities, \reb{we demonstrated a use case for our dataset training models that, given the novelty of the multimodal nature of Pokeflex, constitute the state-of-the-art in multi-object online template-based mesh reconstruction from multimodal data, to the best of our knowledge.} We refer the reader to our \href{https://pokeflex-dataset.github.io/}{website}\footnote{\scriptsize \url{https://pokeflex-dataset.github.io/}} or the supplementary material for further demos and examples.

\end{abstract}

\section{Introduction}

% Motivation 
The development of high-quality datasets is essential to advance research in deformable object manipulation using data-driven methods, which have recently demonstrated promising results in fields such as healthcare, food processing, and manufacturing \cite{orbitSurgical, shi2023robocook, avigal2022speedfolding, bartsch2024SculpBot}.  Such datasets are crucial for training manipulation policies, estimating material parameters, and training 3D mesh reconstruction models. The latter, in particular, plays a vital role in facilitating the close-loop execution of control methods based on mesh simulations \cite{duenser2018benderbot}.
% Objective
In light of these needs, the objective of this work is to create a reproducible, diverse, and high-quality dataset for deformable volumetric objects, grounded in real-world data.

\begin{table*}[h!]
    \centering
    \caption{
    Feature comparison of the PokeFlex dataset with other deformable object datasets.
    }
    \label{tab:comparison}
    % \begin{tabular}{lllllllll}
%                 & {\bf Real-world} &{\bf Meshes} &{\bf Point Clouds} &{\bf RGB Images} &{\bf Force Torque}  &{\bf \# of Objects} &{\bf \# of Frames} &{\bf Type of Deformation}
% \\   \\

% {\bf PokeFlex (ours)}  
% &{\bf \checkmark} 
% &{\bf \checkmark} 
% &{\bf \checkmark} 
% &{\bf \checkmark} 
% &{\bf \checkmark} 
% &\HZ{xxx} 
% &\HZ{xxx} 
% &\HZ{Poke, Drop}\\

% HMDO \cite{XIE2023HMDO_CuteToys} 
% &{\bf \checkmark} 
% &{\bf \checkmark} 
% & 
% &{\bf \checkmark} 
% & 
% &12 
% &2,166 
% &Manual\\

% PLUSH \cite{chen2022virtual}
% &{\bf \checkmark} 
% & 
% &{\bf \checkmark} 
% &{\bf \checkmark} 
% &Force 
% &12 
% &22.84K 
% &Airstream\\

% DOT \cite{Li2024DOT} 
% &{\bf \checkmark} 
% & 
% &{\bf \checkmark} 
% &{\bf \checkmark} 
% & 
% &4 
% &117K 
% &Manual\\

% Household Cloth Object Set 
% &{\bf \checkmark} 
% &{\bf \checkmark}\hyperlink{hyref:01}{$^\dag$} 
% & 
% &{\bf \checkmark} 
% & 
% &27 
% &67
% &Static\\

% Defgraspsim \cite{huang2022defgraspsim} 
% & 
% &{\bf \checkmark}
% &
% &
% &
% &34 
% &1.1M 
% &Grasp \\

% \end{tabular}

\resizebox{0.95\textwidth}{!}{%
\begin{tabular}{@{}lcccccccc@{}}
\toprule
\multicolumn{1}{c}{} &
  \textbf{\begin{tabular}[c]{@{}c@{}}Real-\\ world\end{tabular}} &
  \textbf{Meshes} &
  \textbf{\begin{tabular}[c]{@{}c@{}}Point\\ clouds\end{tabular}} &
  \textbf{\begin{tabular}[c]{@{}c@{}}RGB\\ images\end{tabular}} &
  \textbf{\begin{tabular}[c]{@{}c@{}}Force\\ torque\end{tabular}} &
  \textbf{\begin{tabular}[c]{@{}c@{}}\# of \\ objects\end{tabular}} &
  \textbf{\begin{tabular}[c]{@{}c@{}}\# of \\ time frames\end{tabular}} &
  \textbf{\begin{tabular}[c]{@{}c@{}}Type of \\ deformation\end{tabular}} \\ 
  \midrule
\textbf{PokeFlex (ours)} &
  \textbf{\checkmark} &
  \textbf{\checkmark} &
  \textbf{\checkmark} &
  \textbf{\checkmark} &
  \textbf{\checkmark} &
  18 &
  21.3k & %16094+180*17=19154
  Poke, drop \\  
HMDO \cite{XIE2023HMDO_CuteToys} &
  \textbf{\checkmark} &
  \textbf{\checkmark} &
   &
  \textbf{\checkmark} &
   &
  12 &
  2,166 &
  Manual\hyperlink{hyref:01}{$^\dag$} \\  
PLUSH \cite{chen2022virtual} &
  \textbf{\checkmark} &
   &
  \textbf{\checkmark} &
  \textbf{\checkmark} &
  Force\hyperlink{hyref:02}{$^\ddag$} &
  12 &
  22.84k &
  Airstream \\  
DOT \cite{Li2024DOT} &
  \textbf{\checkmark} &
   &
  \textbf{\checkmark} &
  \textbf{\checkmark} &
   &
  4 &
  117k &
  Manual\\  
  Household Cloth Object Set \cite{garcia2022household} &
  \textbf{\checkmark} &
  \textbf{\checkmark}\hyperlink{hyref:03}{$^\S$}  &
   &
  \textbf{\checkmark} &
   &
  27 &
  67 &
  / \\  
Defgraspsim \cite{huang2022defgraspsim} &
   &
  \textbf{\checkmark} &
   &
   &
   &
  34 &
  1.1M &
  Grasp \\  
  \bottomrule
\end{tabular}%
}

    {\footnotesize
\\
\hypertarget{hyref:01}{$^\dag$} by hand \hspace{2em}
\hypertarget{hyref:02}{$^\ddag$} by providing air nozzle poses
\hspace{2em}
\hypertarget{hyref:03}{$^\S$} for ten static scenes of the cloth objects folded \hspace{3em}
%\hypertarget{hyref:03}{$^\S$} not publicly available as of Oct. 1, 2024
}
\end{table*}

% Challenge
Current state-of-the-art simulation methods offer an attractive alternative to collect such datasets, providing easy access to privileged information such as deformed mesh configurations and contact forces \cite{tripicchio24cepb, huang2022defgraspsim, warp2022, Qiao2021Differentiable, todorov2012mujoco, faure2012sofa}. However, such simulators require careful system identification and fine-tuning to bridge the sim-to-real gap, which ultimately requires real-world data. Static scans rotating around the scene \cite{dinesh2001scanning, garcia2022household} or custom multi-camera systems \cite{chen2022virtual} can be used to collect real-world 3D models. The former can be excessively time-consuming and is unsuitable to capture temporal dynamics. The latter requires careful synchronization and data curation, especially when using noisy lower-cost sensors.

% Approach
To address these challenges, we leverage a professional multi-view volumetric capture system (MVS) that allows capturing detailed 360° mesh reconstructions of deformable objects over time \cite{collet2015MVS}, which we use as ground-truth meshes. We integrate a robotic manipulator with joint-torque sensing capabilities into the MVS, 
enabling contact force estimation and automated data collection. Moreover, to enhance reproducibility and to expand the diversity of data modalities, we also integrate and synchronize lower-cost Azure Kinect and Intel RealSense D405 RGB-D sensors into the MVS.

\begin{figure}[b]
\centering
\includegraphics[width=0.99\linewidth]{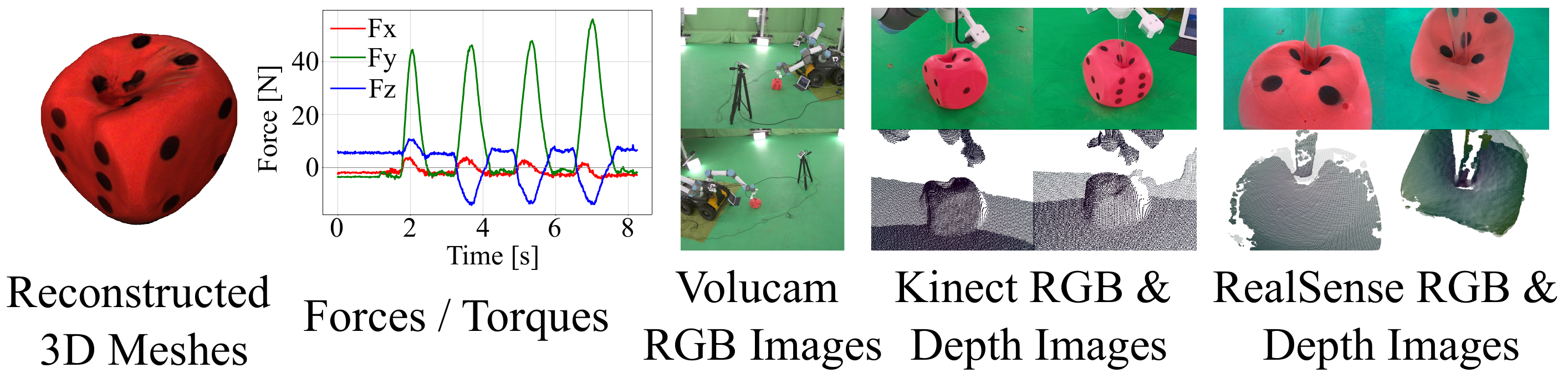}
% \vspace{0.3cm}
\caption{Different data modalities provided by PokeFlex.
}
\label{fig:modalities}
\end{figure}

Our work proposes the PokeFlex dataset (\Cref{fig:overview}), featuring the real-world behavior of 18 deformable objects, including everyday and 3D-printed objects. Deformations are generated via controlled poking and dropping protocols. An overview of the paired, synchronized, and annotated data is illustrated in \Cref{fig:modalities}, and summarized in \Cref{tab:dataset_content}. We demonstrated a use case of the PokeFlex dataset, proposing baseline models capable of ingesting PokeFlex multimodal data and presenting evaluation criteria for benchmarking the results. Specifically, we train neural network models for deformed mesh reconstructions based on template meshes and various input data modalities, including images, point clouds, end-effector poses and forces. The proposed architectures are suitable for online applications, reconstructing 3D meshes at a range from \reb{33}\,Hz to \reb{185}\,Hz depending on the input data modality, on a desktop PC with an NVIDIA RTX 4090 GPU. The pretrained models will be available with PokeFlex.

%capable of reconstructing different object meshes at between 106-215\,Hz, suitable for online applications. 

%We also verify generalizability of the learned models by evaluating them using unseen novel viewpoints of the scene. 

\section{Related work}
\textbf{Deformable object datasets.}  Depending on the use of synthetic or real-world data, deformable object datasets can be roughly categorized into two major groups. \cite{huang2022defgraspsim}, for instance, evaluates multiple grasping poses for deformable objects on a large-scale synthetic dataset. Qualitative sim-to-real experiments for such dataset, show that their simulator captures the general deformation behavior of objects during grasping. 
% Similarly, \cite{lu2024garmentlabunifiedsimulationbenchmark} introduces a simulation environment and benchmark for deformable object and garment manipulation, incorporating static scans of real-world objects to generate simulation models. Notably, they also scan 3 plush toys in static configurations. 
However, careful system identification and parameter tuning are necessary to achieve higher sim-to-real fidelity for synthetic datasets. 

\begin{table}[b]
    \centering
    \caption{Dataset overview (per object, per sequence).}
    \footnotesize 
    \begin{adjustbox}{width=\columnwidth}
\begin{tabular}{lcc}
        \toprule
        
        \textbf{Sequence Data} & \textbf{Poking} & \textbf{Dropping} \\ \midrule
        
        \textbullet\ 3D textured deformed mesh model & \checkmark & \checkmark\\ 
          
        \begin{tabular}[c]{@{}l@{}}\textbullet\  \\ \, \end{tabular} \begin{tabular}[c]{@{}l@{}}RGB images from two Volucam cameras \\ (cameras from the MVS)\end{tabular}&
        \checkmark & \checkmark\\

        \begin{tabular}[c]{@{}l@{}}\textbullet\  \\ \, \end{tabular}
        \begin{tabular}[c]{@{}l@{}}RGB-D images from two Intel RealSense \\D405 sensors (eye-in-hand mounted)\end{tabular}&
        \checkmark \\ 
        
        \begin{tabular}[c]{@{}l@{}}\textbullet\  \\ \, \end{tabular} \begin{tabular}[c]{@{}l@{}}RGB-D images from two Azure Kinect \\ sensors (eye-to-hand mounted)\end{tabular}&
        \checkmark \\ 
        
        \textbullet\ Estimated 3D contact forces and torques & \checkmark \\ 
       \textbullet\  End-effector poses \reb{and contact locations} & \checkmark \\
        
        \midrule
        
        \textbf{Camera and Object Data} & \multicolumn{2}{c}{} \\ \midrule
        \textbullet\ Camera intrinsic and extrinsic parameters &\multicolumn{2}{c}{\checkmark} \\ 
        \textbullet\ 3D textured template mesh model &\multicolumn{2}{c}{\checkmark}\\ 
        
        \begin{tabular}[c]{@{}l@{}}\textbullet\  \\ \, \end{tabular} \begin{tabular}[c]{@{}l@{}}Open-source print files to reproduce \\ the 3D printed objects\end{tabular}&
        \multicolumn{2}{c}{\checkmark} \\ 

        \bottomrule
    \end{tabular}
\end{adjustbox}
    \label{tab:dataset_content}
\end{table}

On the other hand, real-world data collection opens up the door to better capture the complex behavior of deformable objects. Current real-world datasets focus mostly on RGB images. HMDO \cite{XIE2023HMDO_CuteToys} also provides real-world 3D meshes for objects undergoing deformation due to hand manipulation. However, they fell short of providing point cloud or force contact information. \cite{chen2022virtual}  provides points clouds and force contact information but it does not perform 3D mesh reconstruction and the deformations are only globally produced using an airstream. \cite{Li2024DOT} offer a large number of frames, however, the object diversity in their dataset is limited. 
% \cite{zhang2024dofsrealworld3ddeformable} presents a pilot dataset with only one type of deformable object under quasi-static deformation, limited camera views, and no reported interaction forces.

In a departure from other datasets, PokeFlex offers a more comprehensive list of features including; 3D meshes, point clouds, contact forces, higher diversity of objects, and multiple types of deformations as detailed in \Cref{tab:comparison}. For simplicity, we report only the effective number of paired time frames in our table, in contrast to what is reported by \cite{XIE2023HMDO_CuteToys} and \cite{Li2024DOT}, where the total number of samples is computed as the number of time-frames times the number of cameras. 

% \cite{dinesh2001scanning} related on deformable objects.

\begin{figure*}[t!]%
    \centering
    \subfloat[]{ \includegraphics[width=0.45\linewidth]{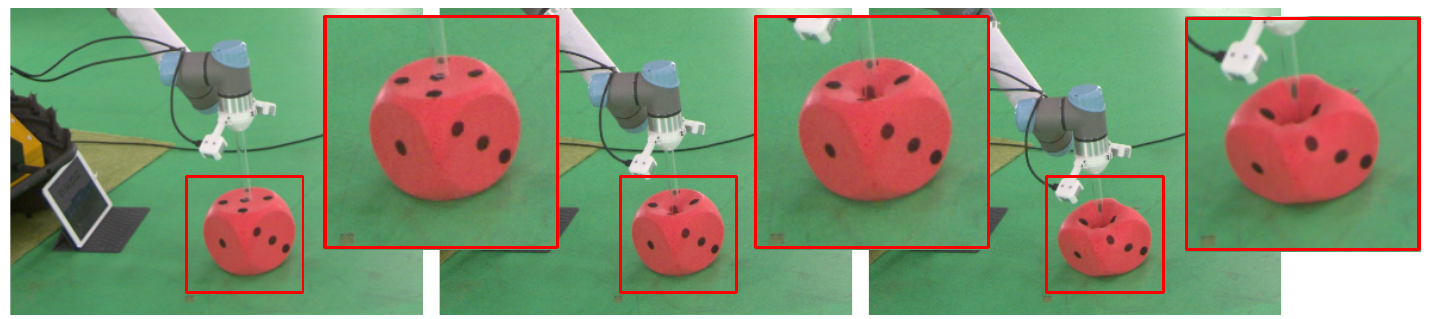} \label{fig:poking_sequence} }
    \hfill
    \subfloat[]{ \includegraphics[width=0.45\linewidth]{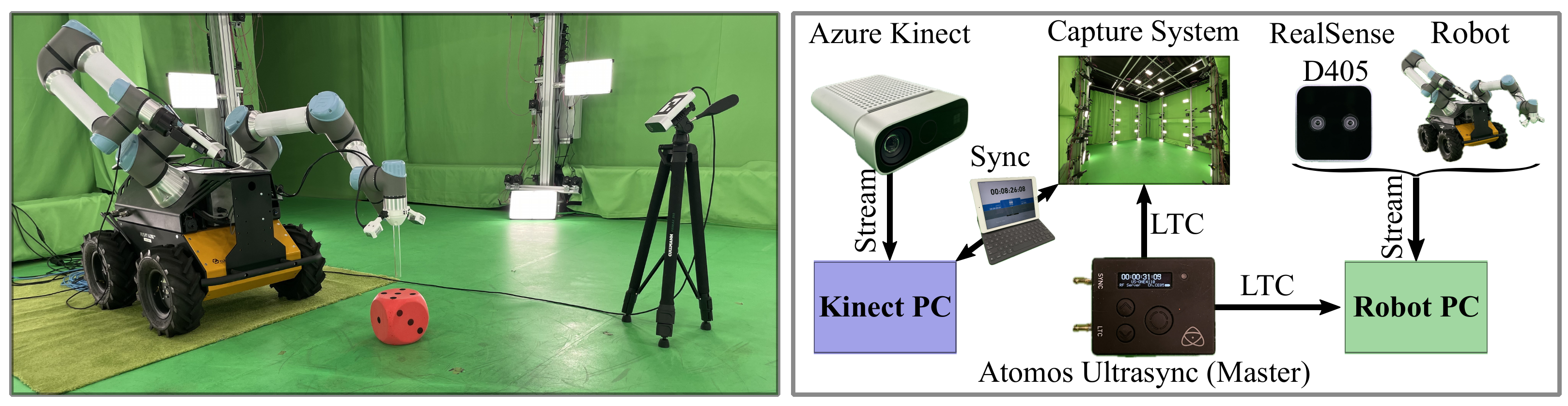} \label{fig:setup} }%
    \caption{a) Sample frames from a poking sequence, with a close-up onto the foam dice. b) \textbf{Left}:  Robotic manipulator positioned inside MVS with external lower-cost cameras sensors. \textbf{Right}: Overview of the system architecture to capture PokeFlex data.}%
    %\label{fig:example}%
    \vspace{-5mm}
\end{figure*}

% \begin{figure*}[t!]
%     %% \centering
%     \begin{minipage}{0.48\textwidth}
%         \centering
%         \includegraphics[width=\linewidth]{images/robot_seq.pdf}
%         \caption{Sample frames from a poking sequence, with a close-up onto the foam dice.}
%         \label{fig:poking_sequence}
%     \end{minipage}%
%     \hfill
%     \begin{minipage}{0.48\textwidth}
%         \centering
%         \includegraphics[width=\linewidth]{images/setup.pdf}
%         \caption{\textbf{Left}:  Robotic manipulator positioned inside MVS with external lower-cost camera sensors during a poking sequence. \textbf{Right}: Overview of the system architecture to capture PokeFlex data.}
%         \label{fig:setup}
%     \end{minipage}
% \end{figure*}

\textbf{Data-driven mesh reconstruction methods} vary widely in terms of the input data modalities they employ. Previous approaches that rely on point clouds to predict deformations are typically trained on synthetic data \cite{mansour2024fast, Lei_2022_CVPR, Niemeyer_2019_ICCV}. While synthetic training data offers controlled and dense point cloud representations, it often leads to a sim-to-real gap as real-world point cloud measurements tend to be noisy and sparse, especially in dynamic and unstructured environments. %This discrepancy can significantly impact the performance and robustness of mesh reconstruction methods when applied to real-world scenarios.
In contrast, methods using single images as input have gained attention for their real-world reconstruction capability without the need for depth information \cite{wang2021pixel, jack2018learning, kanazawa2018learning}. 
However, many of these image-based approaches are not optimized for online inference, making them unsuitable for downstream applications in robotics, where online feedback is essential. For instance, \cite{xu2024instantmeshefficient3dmesh} proposes an instant image-to-3D framework to generate high-quality 3D assets, but requires up to 10 seconds per frame, limiting its practicality for scenarios demanding real-time processing.

% \begin{figure*}[t!]
%     %% \centering
%     \begin{minipage}{0.45\textwidth}
%         \centering
%         \includegraphics[width=0.9\linewidth]{images/robot_seq.pdf}
%         \caption{Sample frames from a poking sequence, with a close-up onto the foam dice.}
%         \label{fig:poking_sequence}
%     \end{minipage}%
%     \hfill
%     \begin{minipage}{0.45\textwidth}
%         \centering
%         \includegraphics[width=0.9\linewidth]{images/setup.pdf}
%         \caption{\textbf{Left}:  Robotic manipulator positioned inside MVS with external lower-cost camera sensors during a poking sequence. \textbf{Right}: Overview of the system architecture to capture PokeFlex data.}
%         \label{fig:setup}
%     \end{minipage}
% \end{figure*}

\section{Methodology}

\subsection{Data Acquisition}

The PokeFlex dataset involves the acquisition of deformations under two different protocols (i) poking and (ii) dropping.
In the poking protocol, a robotic manipulator pokes objects with a transparent acrylic stick multiple times along a randomly oriented horizontal vector (\Cref{fig:poking_sequence}). The dataset also provides the CAD model for the mounting tool, which holds two RealSense cameras and a 192 mm long acrylic stick with a radius of 10 mm. In the dropping protocol, objects are attached to a light nylon cord at approximately 2\,m height and captured in a free-fall drop onto a flat surface. We record data at 30 fps and 60 fps for the poking and dropping protocols, respectively. We leverage a professional multi-view volumetric capture system (MVS), consisting of 106 cameras (53 RGB / 53 infrared) with 12 MP resolution.

% \begin{figure}[t!]
% \centering
% \includegraphics[width=0.99\linewidth]{images/robot_seq.pdf}
% %\vspace{-0.5\baselineskip}
% \caption{Sample frames from a poking sequence, with a close-up onto the foam dice.}
% \vspace{-1.5\baselineskip}
% \label{fig:poking_sequence}
% \end{figure}

% \begin{figure}[t!]
% \centering
% %\vspace{0.3cm}
% \includegraphics[width=0.99\linewidth]{images/setup.pdf}
% \vspace{-0.5\baselineskip}
% \caption{\textbf{Left}:  Robotic manipulator positioned inside MVS with external lower-cost camera sensors during a poking sequence. \textbf{Right}: Overview of the system architecture to capture PokeFlex data.}
% \vspace{-1.0\baselineskip}
% \label{fig:setup}
% \end{figure}

For the poking protocol, we integrated and synchronized additional hardware to the MVS to ensure temporally aligned data capture across all modalities. The additional hardware includes the robot manipulator and four additional RGB-D cameras: two Azure Kinect cameras to capture the scene from opposing viewpoints, and two Intel RealSense D405 cameras mounted on the robot's end-effector. The robot logs end-effector poses, interaction forces and torques at 120\,Hz, while these four cameras record RGB-D data at 30\,Hz.

To synchronize devices, we rely on a Linear Timecode (LTC) signal provided by an Atomos Ultrasync device. The cameras of the MVS have a leader/follower architecture, where the internal clocks of the follower cameras are synchronized to one single leader camera, which reads the LTC signal.
In addition to the MVS control system, we use two desktop PCs to read the additional data streams: a Robot PC that reads the robot data and the streams of the two RealSense D405 cameras and a dedicated Kinect PC that reads the streams of the two Azure Kinect devices.

The Robot PC is synchronized with the capture system by reading the same LTC signal provided by the Atomos Ultrasync device. The Kinect cameras are hardware-synchronized with each other. Their synchronization with the capture system is achieved retrospectively by comparing the current timecode displayed on a screen in the camera frames of the Kinect and the camera frames of the capture system. An overview of the architecture is shown on \Cref{fig:setup} (Right).

We utilize a system similar to that described by \cite{collet2015MVS} to reconstruct the meshes and textures of the objects under deformation. When recording at 30 fps, the MVS generates approximately 27 GB of raw data per second. This data is then processed using commercial software provided by Acturus Studio on 10x On-Prem Nodes servers, achieving an output rate of approximately one 3D frame per minute. The reconstructed meshes and textures were curated to ensure that only the deformable objects were retained in the scene.

%\MZ{ @Juan please help!:  please help us here to write the specs of the server, to properly cite Arcturus and the HoloApp, and to describe the HoloApp a bit more, if necessary. } \JZ{yes, will do on Monday}
%Mesh reconstructions of the captured sequences are computed using the HoloApp software from Arcturus on a server with a \# cpus and \# GPUS. The total computational time invested in mesh reconstruction was 14 days, translating to approximately 1 minute of compute time per frame.

\subsection{Learning-based Mesh Reconstruction}

\begin{figure*}[t!]
\centering
\includegraphics[width=0.8\textwidth]{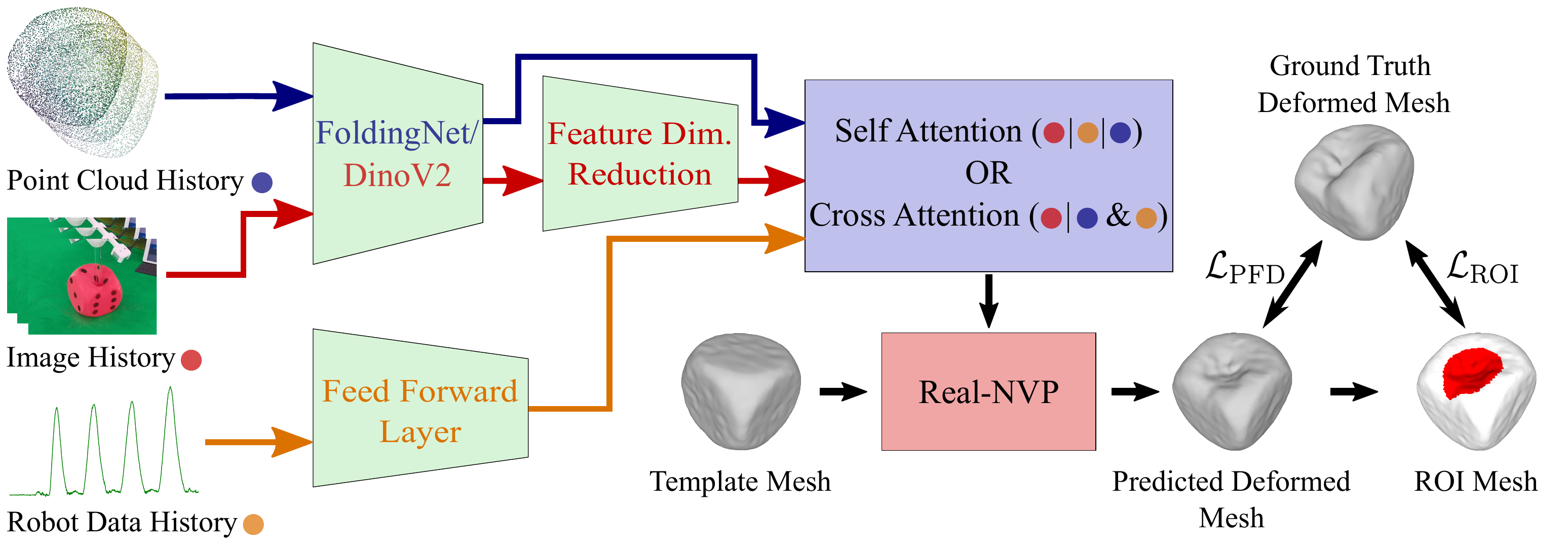}
\caption{Superimposed representation of the proposed network architectures for ingesting the multi-modal PokeFlex data.}
\vspace{-0.5\baselineskip}
\label{fig:image_architecture}
\end{figure*}

We leverage PokeFlex to train models for template-based mesh reconstruction, inferring the deformation of a template mesh from an input of history buffers with various combinations of data modalities: images, point clouds, and/or robot data. \Cref{fig:image_architecture} illustrates the building blocks used to generate different architectures tailored to the data modalities. 

%\begin{wrapfigure}{r}{0.5\linewidth}

At a high level, we use three main common components for all models: an encoder to extract features from an input modality, an attention mechanism to exploit temporal information from the history buffers, and a conditional Real-NVP~\cite{mansour2024fast} to predict the offsets of template vertices, yielding the predicted deformed mesh. Real-NVP uses a series of conditional coupling blocks, defined as continuous bijective functions. This continuous bijective operation ensures that the model is homeomorphic, allowing stable deformation of a template mesh $\mM_\text{T}$, while preserving its topology.
\vspace{3mm}

\textbf{Image input:} For pipelines using images as input, we use a DinoV2 vision transformer to extract embeddings of each image frame. In particular, we use a DinoV2-small model,  pretrained via distillation from the largest DinoV2 transformer presented in \cite{oquab2023dinov2} (LVD-142M dataset). The embedding dimension is later reduced using a 1D convolutional layer and a subsequent fully connected layer (Feature Dim. Reduction block  in \Cref{fig:image_architecture}). %During training we finetune the model with PokeFlex data.

\textbf{Point cloud input:} When using point clouds, we leverage a FoldingNet encoder~\cite{yang2018foldingnet} for representation learning, which is trained end-to-end together with the attention mechanism and the conditional-NVP.  

%\FO{missing details. Is the FoldingNet trained on Pokeflex training set? Is the latent embeddings of foldingnet autoencoder passed directly to self-attention layer?}. 
%\HZ{Yes, foldingnet is trained on pokeflex training set.}
%\MZ{Yes, the latent embeddings of foldingnet autoencoder ared passed directly to self-attention layer}

\textbf{Robot data input:}  To fuse the robot data, we concatenate the measured end-effector forces and the position of the interaction point. The concatenated data is later fed into a single fully connected layer, to match the dimensionality of the embeddings used for the attention mechanisms. 
% \newline

%The model we use for observations consisting of robot manipulator data has a single feed-forward layer to encode the measured forces and the position of the interaction point. 

% \textbf{Attention mechanisms:}
A self-attention mechanism is employed for variations of the architecture in \Cref{fig:image_architecture} that use a single data modality as input. In contrast, a cross-attention mechanism is applied when handling multiple data modalities simultaneously. For the experiments presented in the results section, we use cross-attention to handle a mixture of history buffers of robot data with images or Kinect point clouds, as input. However, other combinations of input data are also possible.

All architectures are end-to-end trained using the same loss. 
%function $\mathcal{L} = \mathcal{L}_{\text{PFD}} + 0.5\,\mathcal{L}_{\text{ROI}}$. 
We include the weights of the DinoV2 transformer during backpropagation for finetuning.
% The main criterion $\mathcal{L}_{\text{PFD}}$ is the bidirectional mean squared point face distance $d(p,f)$, computed between the set of sampled points $p_i \in \mathcal{P}$ of the predicted mesh and the set of triangular faces $f_i \in \mathcal{F}$ of the ground truth mesh, which accounts for the global deformation of the objects (\cref{eq:lpfd}). 
The main point face distance (PFD) criterion $\mathcal{L}_{\text{PFD}}$ accounts for the global deformation of the objects, which computes the average squared distance $d(\vp,\vf)$ from the set of sampled points $\vp_i \in \mathcal{P}$ on the predicted mesh to the nearest faces in the set of triangular faces $\vf_i \in \mathcal{F}$ of the ground truth mesh and vice versa (\cref{eq:lpfd}).
% Moreover, to deal with the local deformations generated in the poking region, we add a loss term $\mathcal{L}_{\text{ROI}}$ (\cref{eq:lroi}) that computes the unidirectional chamfer distance between the set of sampled points $q_i \in \mathcal{Q}$ of the ground truth mesh and the points $p_i$ in the region of interest. 
Moreover, to deal with the local deformations generated in the poking region, we add a region-of-interest (ROI) loss $\mathcal{L}_{\text{ROI}}$ (\cref{eq:lroi}) that computes the unidirectional chamfer distance from the points $\vp_i$ in the ROI to the set of sampled points $\vq_i \in \mathcal{Q}$ of the ground truth mesh. 
The ROI is defined using the indicator function $\mathbb{I}(\mathcal{C}(\vp_i))$, which evaluates to 1 if point $\vp_i$ is close enough to the contact point $\vt$ according to a threshold $\epsilon$, and if the minimum vertical component of the contact point $\vp_{i,y}$ is bigger than a vertical threshold $\epsilon_{y}$ (\cref{eq:cpi}).
%the minimum vertical coordinate across all the vertices $y_{\min}$ scaled by a factor 

\begin{equation}
\label{eq:lpfd}
\mathcal{L}_{\text{PFD}} = \frac{1}{|\mathcal{P}|} \sum_{\vp_i \in \mathcal{P}} \min_{\vf_j \in \mathcal{F}} d(\vp_i, \vf_j) + \frac{1}{|\mathcal{F}|} \sum_{\vf_j \in \mathcal{F}} \min_{\vp_i \in \mathcal{P}} d(\vf_j, \vp_i) \text{ ,}
\end{equation}

\begin{equation}
\label{eq:lroi}
    \mathcal{L}_{\text{ROI}} = \frac{1}{|\mathcal{P}|} \sum_{\vp_i \in \mathcal{P}} \mathbb{I}(\mathcal{C}(\vp_i)) \cdot \min_{\vq_j \in \mathcal{Q}} \| \vp_i - \vq_j \|^2 \text{ ,}
\end{equation}

\begin{equation}
\label{eq:cpi}
    \mathcal{C}(\vp_i) = \left( \| \vp_i - \vt \| \leq \epsilon \right) \land \left( \vp_{i,y} > \epsilon_{y} \right) \text{ .}
\end{equation}

The total loss is then set as $\mathcal{L} = \mathcal{L}_{\text{PFD}} + 0.5\,\mathcal{L}_{\text{ROI}}$.

% \begin{equation}
% \label{eq:lpfd}
% \mathcal{L}_{\text{PFD}} = \frac{1}{N_p} \sum_{p_i \in \mathcal{P}} \min_{f_j \in \mathcal{F}} d(p_i, f_j) + \frac{1}{N_f} \sum_{f_j \in \mathcal{F}} \min_{p_i \in \mathcal{P}} d(f_j, p_i),
% \end{equation}

% \begin{equation}
% \label{eq:lroi}
%     \mathcal{L}_{\text{ROI}} = \frac{1}{N_p} \sum_{p_i \in \mathcal{P}} \mathbb{I}(\mathcal{C}(p_i)) \cdot \min_{q_j \in \mathcal{Q}} \| p_i - q_j \|^2,
% \end{equation}

% \begin{equation}
% \label{eq:cpi}
%     \mathcal{C}(p_i) = \left( \| p_i - \vt \| \leq r \right) \land \left( p_{i,y} > 0.2 \cdot y_{\min} \right).
% \end{equation}

% $N_p$ is the number of sampled points in $\mathcal{P}$, $N_f$ is the number of faces in $\mathcal{F}$.

% $\textbf{t}$ is the position of the end effector, $r$ is a defined threshold.

\section{Results}
\label{sec:results}
\subsection{Dataset}
\label{sec:results_dataset}

\begin{figure*}[t!]
\centering
\includegraphics[width=0.99\textwidth]{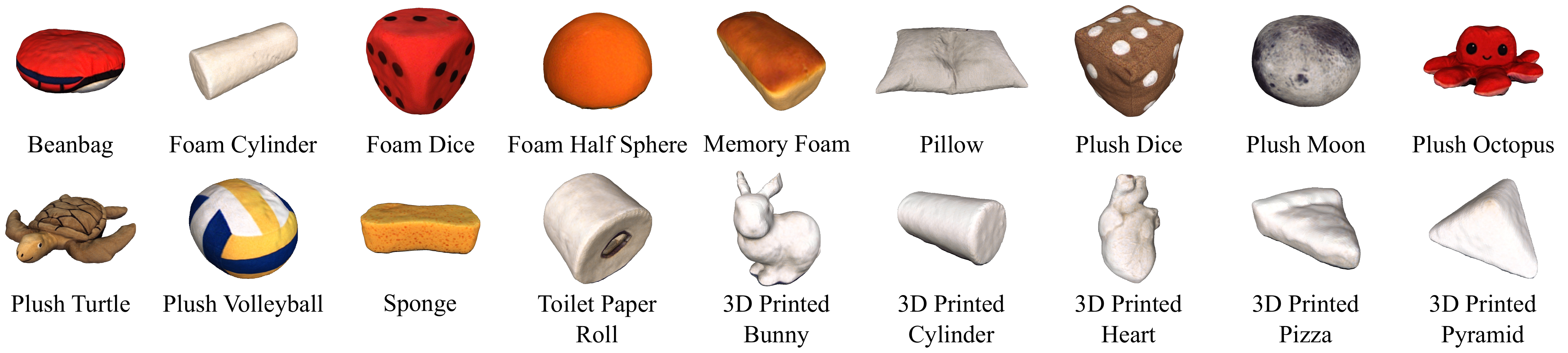}
% \vspace{0.3cm}
\caption{Rest-state reconstructed 3D meshes of all 18 objects featured in the PokeFlex dataset.}
\vspace{-2mm}
\label{fig:objects}
\end{figure*}

% objects
The PokeFlex dataset comprises 18 deformable objects (\Cref{fig:objects}), including 13 everyday items as well as 5 objects that are 3D printed with a soft thermoplastic polyurethane filament. 
% 3d print
Even though the everyday objects in our dataset can be purchased from global vendors, their availability is not guaranteed worldwide. Therefore, to enhance the usability of our dataset we include deformable 3D printed objects, providing print files and detailed specifications for reproducibility. 
% This further allows one to know the exact internal structure of these deformable objects, e.g., useful for a first step for sim-to-real testing.
The 3D  printed objects include the Stanford bunny \cite{Turk1994bunny}, a cylinder, a heart \cite{noor20193dheart}, a pyramid, and a custom pizza slice. \Cref{appx:3dprint} reports further details for 3D printing. %Further details about the 3D printing can be found in \Cref{appx:3dprint}.

% object properties
The dimensions and the weights of the PokeFlex objects range from 7\,cm to 58\,cm and from 22\,g to 1\,kg, respectively. Furthermore, using Hooke's law and applying RANSAC for linear regression to avoid outliers, we estimated the objects' stiffnesses to be in the range of 148–2,156\,N/m.
% (\Cref{fig:stiffness}). 

\begin{figure}[b!]
\centering
\vspace{0.3cm}
\includegraphics[width=0.99\linewidth]{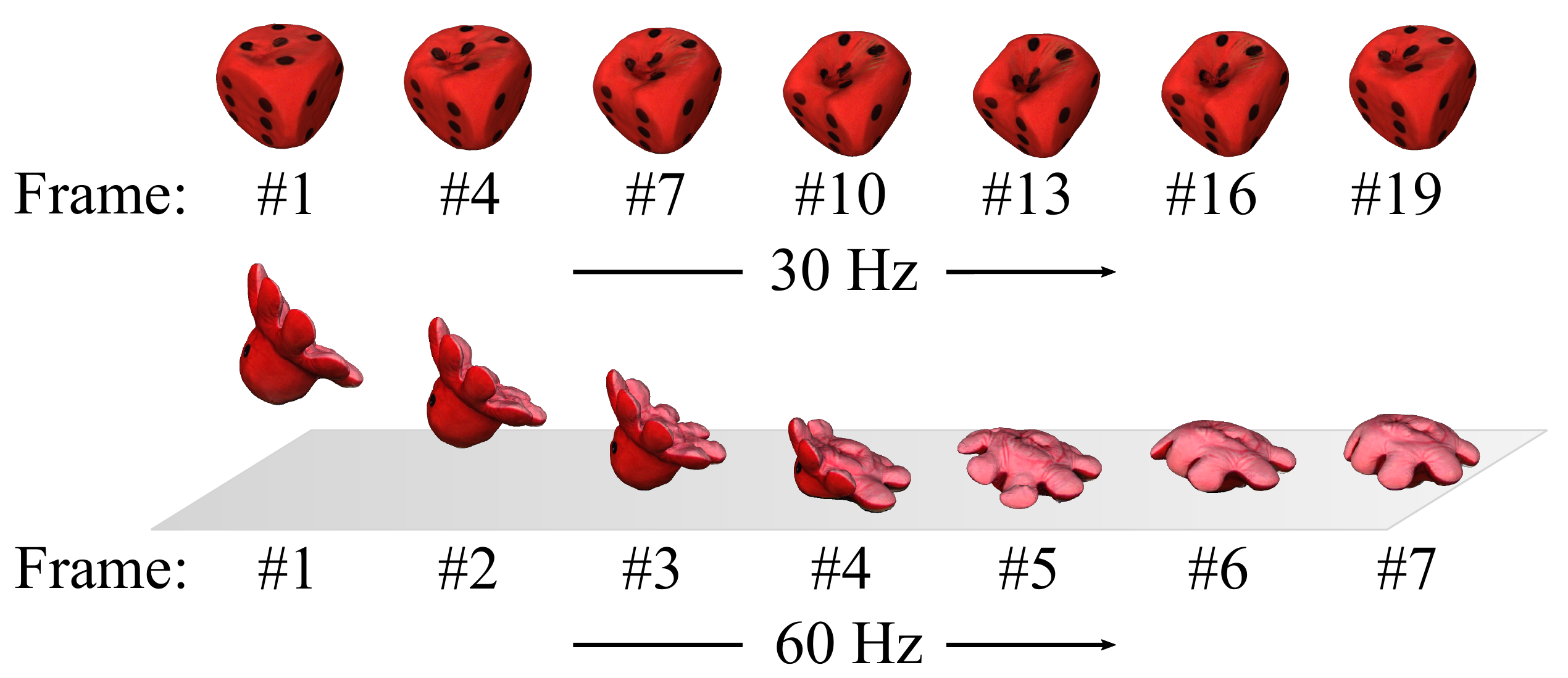}
% \vspace{0.3cm}
         \caption{\textbf{Top:} Mesh reconstructions of foam dice for a poking sequence shown in every third frame. \textbf{Bottom:} Mesh reconstructions of plush octopus for a dropping sequence.} 
% Reconstructed meshes of foam dice in a poking sequence showing every third frame at 30 Hz.}
\label{fig:poking_dropping_sequence}
\end{figure}

% capture
For the poking protocol, we recorded 4-8 sequences with a duration of 5-6 seconds at 30 fps for each object. Similarly, for the dropping protocol, we recorded 3 sequences of 1 second at 60 fps for each object. \Cref{fig:poking_dropping_sequence} shows two reconstructed sequences for poking and dropping. In the case of the poking sequences, each frame includes synchronized and paired data from all modalities, as illustrated in \Cref{fig:modalities}. 

%\HZ{put total number of frames and total number of deformed frames here. 20,014 16,774 3,240}
The total number of reconstructed frames used to generate ground-truth data was \reb{21.3k, which comprises 18.1k frames for the poking sequences and 3.2k frames for the dropping sequences. It is worth noting that after curating the frames of the poking sequences, i.e., discarding the frames where the robot arm is not in contact with the objects, the total number of active paired poking frames reduces from 18.1k down to 8.6k. Furthermore, if we consider all the different modalities to count the number of samples as done in \cite{Li2024DOT, XIE2023HMDO_CuteToys}, the total number of samples in PokeFlex amounts to 240k. 
}  %8,109
% summary
 A summary of the physical properties of the objects, as well as a per-object list of the recorded frames under deformation for the poking sequences, is presented in \Cref{appx:object}. For the dropping protocol, we recorded 180 frames per object.

\subsection{Evaluation of Learning-based Reconstruction}
\label{sec:results_reconstruction}
% Roadmap: 
% 0. Talk about the data
% 1. We report additional metrics. 
% 2. First experiment is the only one that uses singles image: dice.
%    \textbf{Table2. Single source, image sequence - single object - different cameras:} \\
% 3. All the other image experiments are trained on 5 objects; dice, sponge, plush moon, half sphere, octopus
% \textbf{Table3. Single source - 5 objects - Different modalities:} \\
% 4. Show reconstruction of different quality levels for the image based case.

% Employing the PokeFlex dataset, we introduce a series of benchmarks for different template-based reconstruction applications. To achieve this, we use the data from the poking sequences and aim to predict the underlying deformation using the architectures from \Cref{architecture}. 
\textbf{Overview of training data.}
In the following experiments, we exclusively used poking sequences from the dataset because of the higher diversity of input data modalities available. The size of the history buffer was ablated and set to 5 for better performance. 
%For simplicity, models were trained using either 1 or 5 objects from the dataset. 
% However, no fundamental reason prevents users of the PokeFlex dataset to include all the objects for training. 
%There are, however, no obstacles in PokeFlex that would prevent users to include more or all objects in their training setups.
The train-validation split was generated by randomly choosing one recording sequence per object as the validation set. Additional training hyper- parameters are reported in \Cref{appx:train_detail}.

% Depending on the amount of recordings of the object, the validation samples constitute to 12.5-25\% of all samples from the object. 

%2.9k for 5 objects. 
%738 for dice object.
\textbf{Metrics.} During training, we reposition and re-scale all meshes into a cube of unit size ($[-0.5, 0.5]^3$) to maintain a consistent scale across all objects.
The losses $\mathcal{L}_{\text{PFD}}$ and $\mathcal{L}_{\text{ROI}}$ are computed in this normalized scale. 
%Additionally, we calculate the relative point-to-face distance (RPFD) by dividing $\mathcal{L}_{\text{PFD}}$ by the average point-to-face distance between the template mesh $\mM_\text{T}$ and the ground truth mesh $\mM_{\text{GT}}$. 
%An RPFD value below 1 indicates that the predicted deformed mesh $\mM_\text{P}$ is closer to the ground truth than the undeformed template, with smaller values indicating better accuracy. 
To further assess the prediction accuracy, we evaluate two additional metrics between the predicted mesh $\mM_\text{P}$  and the ground truth mesh  $\mM_{\text{GT}}$ in their original scale: the unidirectional L1 Norm Chamfer Distance $\text{CD}_{\text{UL1}}$ (\cref{eq:cdul1}) and the volumetric Jaccard Index $J$ (\cref{eq:jindex}), which we defined in terms of the volume operator $V$. The metrics provide insights into the L1 Norm surface distance and the volume overlap ratio, respectively. 

\begin{equation}
\label{eq:cdul1}
    \text{CD}_{\text{UL1}} = \frac{1}{|\mathcal{P}|} \sum_{\vp_i \in \mathcal{P}} \min_{\vq_j \in \mathcal{Q}} \| \vp_i - \vq_j \|_1 \text{ ,}
\end{equation}
\begin{equation}
\label{eq:jindex}
J(\mM_{\text{A}}, \mM_{\text{B}}) = \frac{V\left( \mM_{\text{A}} \cap \mM_{\text{B}} \right)}{V\left( \mM_{\text{A}} \cup \mM_{\text{B}} \right)} \text{ .}
\end{equation}

\begin{table}[b]
    \centering
    \caption{\reb{Mean prediction performance for proposed model configurations trained on all objects. Arrows indicate that a better performance is either higher $\uparrow$ or lower $\downarrow$.     
    }}
    \footnotesize % or use \footnotesize for even smaller text
    \begin{adjustbox}{width=\columnwidth}
        \begin{tabular}{lccccc} % Input % LPFD % LROI % J % RPFD
        \toprule
        \textbf{Input} & 
        \begin{tabular}[c]{@{}c@{}}\textbf{$\mathcal{L}_{\text{PFD}}\downarrow$} \\ \textbf{$\cdot 10^3 $}\end{tabular}&
        \begin{tabular}[c]{@{}c@{}}\textbf{$\mathcal{L}_{\text{ROI}}\downarrow$} \\ \textbf{$\cdot 10^3 $}\end{tabular}&

        \begin{tabular}[c]{@{}c@{}}\textbf{$\text{CD}_{\text{UL1}} \downarrow$} \\ \textbf{$\text{[mm]}$}\end{tabular}&

        % \textbf{$\mathcal{L}_{\text{ROI}} \cdot 10^3 \downarrow$} 
        % & \textbf{$\text{RPFD} \downarrow$} 
        % \textbf{$\text{CD}_{\text{UL1}} \text{[mm]} \downarrow$}  & 
        \textbf{$J(\mM_{\text{P}}, \mM_{\text{GT}}) \uparrow$}   \\ \midrule
        \textbullet\ Images
        & 7.34 %$\pm$ 13.45          
        & 7.22 %$\pm$ 12.09 
        % & 0.708 $\pm$ 0.783
        & 7.547 %$\pm$ 3.684         
        & 0.791 %$\pm$ 0.126 
        \\ %\hline
        \textbullet\ Robot data
        & 8.47 %$\pm$ 13.57          
        & 5.50 %$\pm$ 9.31
        % & 0.867 $\pm$ 0.934
        & 8.283 %$\pm$ 3.719
        & 0.779 %$\pm$ 0.111  
        \\ %\hline
        \textbullet\ Images + robot data 
        & 6.64 %$\pm$ 12.72          
        & 4.99 %$\pm$ 10.58  
        % & 0.638 $\pm$ 0.716        
        & 7.025 %$\pm$ 2.995          
        & 0.806 %$\pm$ 0.110
        \\ %\hline
        
        \begin{tabular}[c]{@{}l@{}}\textbullet\  \\ \, \end{tabular} \begin{tabular}[c]{@{}l@{}}Dense synthetic point \\ clouds (5k points)\end{tabular}
        & \textbf{4.34} %$\pm$ 5.32
        & \textbf{3.64} %$\pm$ 6.12
        % & 0.621 $\pm$ 0.832
        & \textbf{6.185} %$\pm$ 2.409
        & \textbf{0.825} %$\pm$ 0.105 
        \\
        \begin{tabular}[c]{@{}l@{}}\textbullet\  \\ \, \end{tabular} \begin{tabular}[c]{@{}l@{}}Sparse synthetic point \\ clouds (100 points)\end{tabular} 
        & 4.84 %$\pm$ 6.22 
        & 4.77 %$\pm$ 7.52
        % & 0.611 $\pm$ 0.737
        & 6.445 %$\pm$ 2.321
        & 0.823 %$\pm$ 0.090 
        \\
        \textbullet\ Kinect point clouds
        & 5.75 %$\pm$ 10.54        
        & 5.00 %$\pm$ 8.05   
        % & \textbf{0.597} $\pm$ 0.681   
        & 6.498 %$\pm$ 2.59 
        & 0.820 %$\pm$ 0.112 
        \\ %\hline
        \begin{tabular}[c]{@{}l@{}}\textbullet\  \\ \, \end{tabular} \begin{tabular}[c]{@{}l@{}}Kinect point clouds \\+ robot data \end{tabular}
        & 5.68 %$\pm$ 9.43      
        & 4.06 %$\pm$ 6.86  
        % & 0.617 $\pm$ 0.728
        & 6.615 %$\pm$ 2.871
        & 0.817 %$\pm$ 0.115 
        \\ %\hline
% \begin{tabular}[c]{@{}l@{}}Sparse synthetic Point \\ Clouds (100 points)\end{tabular} & 3.72          & 5.30          & 0.640          & 5.513          & 0.851 \\ 
        \bottomrule
    \end{tabular}
\end{adjustbox}
    \label{tab:performance_metric_diff_data}
\end{table}

\begin{figure}[h]    
\centering
\includegraphics[width=0.99\linewidth]{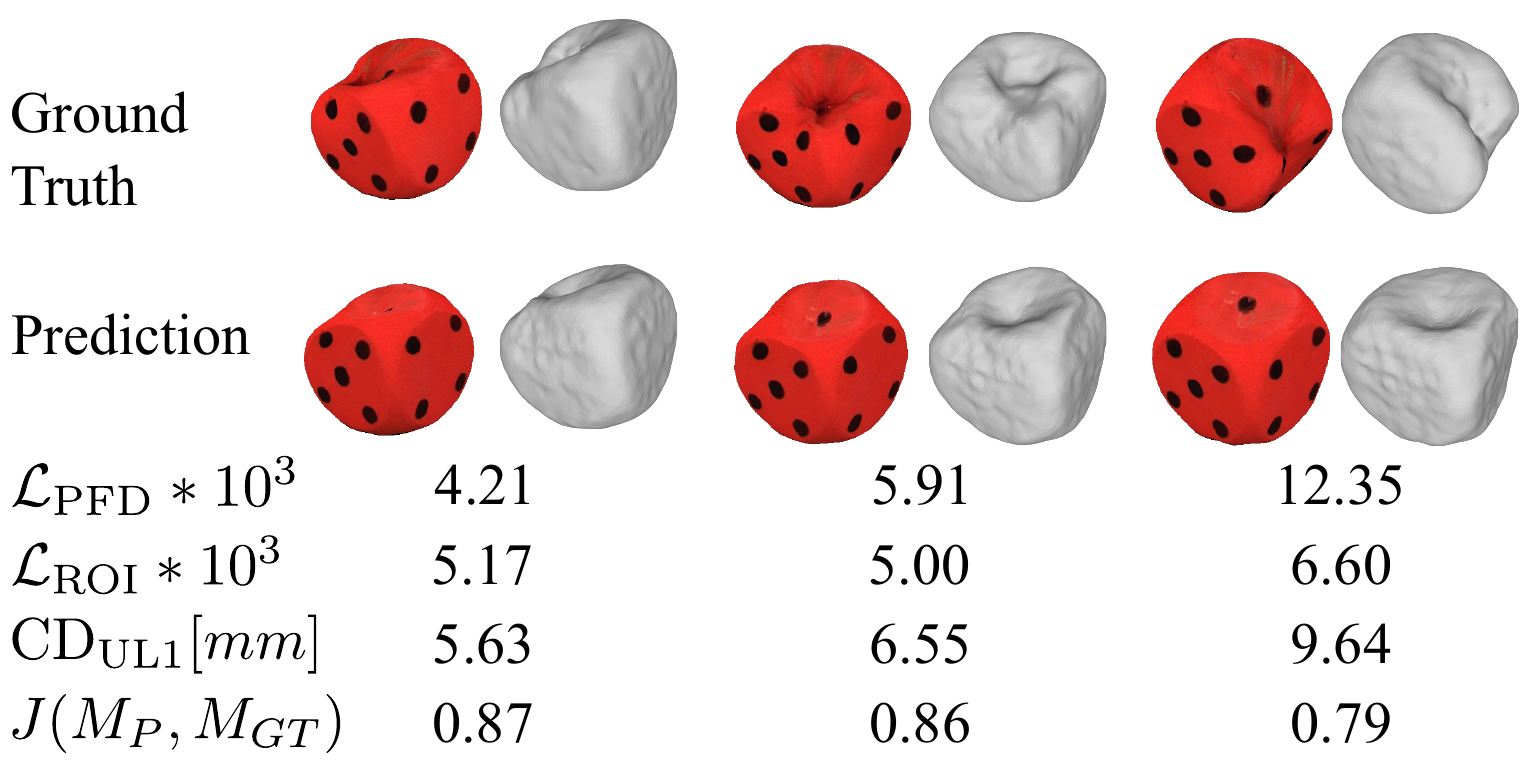}
%\vspace{-0.5\baselineskip}
\caption{Examples of deformation predictions for a foam dice and their corresponding metrics. Meshes are rendered side by side with and without texture to highlight the  ROI.}
%\vspace{-1.0\baselineskip}
\label{fig:predictions}
\end{figure}

\textbf{Learning from different data modalities.}
In this experiment, we train different mesh prediction models from sequences of different input modalities.
% Same as in the previous experiment, w
We trained multi-object models using all 18 objects from the dataset. Detailed performance for the evaluated data modalities can be found in \Cref{tab:performance_metric_diff_data}. Inference rates across different data modalities, detailed in \Cref{appx:inference}, range from \reb{33} Hz to \reb{185} Hz for dense point clouds and forces, respectively. \Cref{fig:predictions} shows examples of predicted meshes with different levels of reconstruction quality obtained using a multi-object model trained from image-sequences only. 
% Additionally, \Cref{appx:object_analytics} reports a detailed breakdown of the per-object performance for models trained from sequences of images, images + robot data, and point clouds. 

\textbf{Generalizing to unseen objects.}
\label{appx:generalization}
To test the generalization capabilities of the proposed architectures, we trained a model on synthetic point cloud data (\Cref{tab:generalization_pcd}) on 14 different objects and evaluated on 4 unseen objects. The generalization performance was evaluated using all the available sequences from the unseen objects.

%and the combination of images and robot data (\Cref{tab:generalization_img_robot_data})

\begin{table}[h]
  \centering
  %\begin{minipage}{0.95\linewidth}
  %\centering
    \caption{Generalization results for 4 unseen objects for point-cloud-based mesh reconstruction.
    }
    \begin{adjustbox}{width=\columnwidth}
\begin{tabular}{lccccc} % Input % LPFD % LROI % J % RPFD
\toprule
\textbf{Input} & 
        \begin{tabular}[c]{@{}c@{}}\textbf{$\mathcal{L}_{\text{PFD}}\downarrow$} \\ \textbf{$\cdot 10^3 $}\end{tabular}&
        \begin{tabular}[c]{@{}c@{}}\textbf{$\mathcal{L}_{\text{ROI}}\downarrow$} \\ \textbf{$\cdot 10^3 $}\end{tabular}&

        \begin{tabular}[c]{@{}c@{}}\textbf{$\text{CD}_{\text{UL1}} \downarrow$} \\ \textbf{$\text{[mm]}$}\end{tabular}&

        % \textbf{$\mathcal{L}_{\text{ROI}} \cdot 10^3 \downarrow$} 
        % & \textbf{$\text{RPFD} \downarrow$} 
        % \textbf{$\text{CD}_{\text{UL1}} \text{[mm]} \downarrow$}  & 
        \textbf{$J(\mM_{\text{P}}, \mM_{\text{GT}}) \uparrow$}   \\ \midrule

\begin{tabular}[c]{@{}l@{}}\textbullet\  \\ \, \end{tabular} \begin{tabular}[c]{@{}l@{}}Validation set \\(14 objects)\end{tabular}
& 5.13 %$\pm$ 6.61          
& 4.26 %$\pm$ 7.85
% & 0.648 $\pm$ 0.742
& 6.247 %$\pm$ 2.731          
& 0.816 %$\pm$ 0.105 
\\ %\hline
\textbullet\ Foam cylinder
& 4.77 %$\pm$ 6.55          
& 2.42 %$\pm$ 1.67
% & 0.629 $\pm$ 0.674
& 8.237 %$\pm$ 3.928          
& 0.806 %$\pm$ 0.093 
\\ %\hline
\textbullet\ Plush dice
& 2.94 %$\pm$ 4.50          
& 3.59 %$\pm$ 4.17
% & 0.571 $\pm$ 0.638
& 7.838 %$\pm$ 1.905          
& 0.895 %$\pm$ 0.036 
\\ %\hline
\textbullet\ Plush volleyball
& 2.48 %$\pm$ 1.93          
& 3.52 %$\pm$ 3.87
% & 0.175 $\pm$ 0.289
& 6.074 %$\pm$ 1.510          
& 0.903 %$\pm$ 0.034 
\\ %\hline
\textbullet\ Sponge
& 9.70 %$\pm$ 9.13          
& 3.54 %$\pm$ 2.31
% & 0.718 $\pm$ 0.764
& 8.673 %$\pm$ 3.42          
& 0.711 %$\pm$ 0.09 
\\ %\hline
    \bottomrule
\end{tabular}
\end{adjustbox}
    \label{tab:generalization_pcd}
    %\end{minipage}
\end{table}

% \begin{table}[h]
%   \centering
%   %\begin{minipage}{0.95\linewidth}
%   %\centering
%     \caption{Generalization results for 4 unseen objects using images and robot data as input.
%     }
%     \input{tables/table-generalization-img-robot-data}
%     \label{tab:generalization_img_robot_data}
%     %\end{minipage}
% \end{table}
% \newpage
\section{Discussion}

% Roadmap:
% 0. Discuss quality of ground-truth data.
% 1. Discuss quality of recorded forces with poking and estimated stiffness. 
% 2. Discuss different cameras.
% 3. Discuss image-based mesh reconstruction trained on multiple objects
% 3.5 Discuss overfitting. 
% 4. Discuss other modalities.

%Using the volumetric capture system we are capable of reconstructing textured meshes of deformed objects when being poked or hitting the floor upon being dropped. 
%Moreover, the PokeFlex dataset offers a range of data modalities capturing the deformation, allowing for the development of various downstream applications.  

% \begin{wrapfigure}{r}{0.4\linewidth}
% \includegraphics[width=\linewidth]{images/memorization.pdf}
% \vspace{-8mm}
% \caption{Assessment of overfitting of our prediction methodology for image based mesh reconstruction.  $\mathcal{L}_{\text{PFD}}$ from the predicted mesh using the ground truth image is normalized with $\mathcal{L}_{\text{PFD}}$ from the predicted mesh using the nearest-neighbour image from the training set.}
% \vspace{-25mm}
% \label{fig:memorization}
% \end{wrapfigure}

\begin{figure}[h!]%{0.5\linewidth}
\centering
\includegraphics[width=0.99\linewidth]{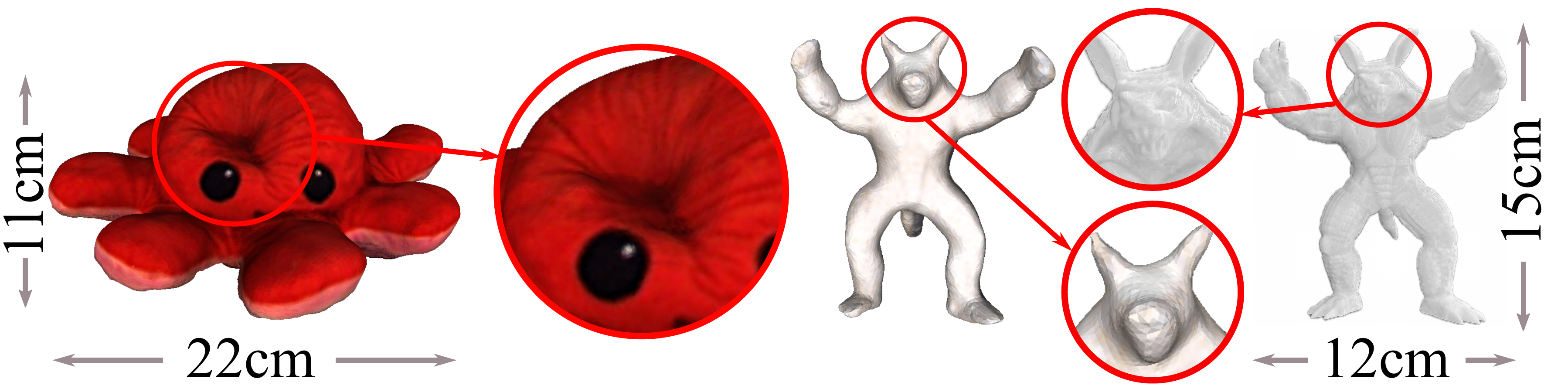}
\caption{Examples of reconstructed ground truth meshes for medium (\textbf{Left}) and small (\textbf{Right}) size objects in deformed states. Reconstruction of fine-grained details is a limitation of our current setup (close-up views on the Right).}
\label{fig:reconstruct_limit}
\end{figure}

\textbf{Quality of ground-truth meshes.} The overall geometry of the objects in the dataset, in static configurations, is well captured by the meshes reconstructed with the MVS as shown in \Cref{fig:objects}, even though the system's intended use is the reconstruction of human-size objects. Furthermore, the proposed poking protocol, using a transparent acrylic stick, helps prevent occlusions at the contact point, leading to detailed reconstruction of objects even when they undergo deformations, as can be seen in \Cref{fig:reconstruct_limit} (Left). However, reconstruction of fine-grained details for smaller objects such as the 3D-printed Stanford armadillo \cite{CurlessLevoy1996armadillo} remains challenging with the current setup of the MVS (\Cref{fig:reconstruct_limit} - Right). Better fine-grained reconstruction results can be expected by rearranging the cameras in a smaller workspace. \newline

\textbf{Multi-object mesh reconstruction from different modalities. } \Cref{tab:performance_metric_diff_data} shows that the dense synthetic point clouds yield the best performance among all data modalities. A drop in performance is observed for the sparser synthetic point clouds, and the noisier point clouds captured by the Kinect. The models trained from robot data and images or Kinect point clouds generally outperformed the counterparts trained from only robot data, only images, or only Kinect point clouds, showcasing the importance of combining different input modalities, and hinting at the effectiveness of our cross-attention mechanism.  %Combining robot data with the Kinect point clouds also leads to performance improvements relative to only using the Kinect point clouds, however the performance gains are more subtle.

%Specifically, the dense synthetic point clouds perform best in terms of validation losses and RPFD, while Kinect point clouds perform best in terms of $\text{CD}_{\text{UL1}}$ and $J(\mM_{\text{P}}, \mM_{\text{GT}})$. Furthermore, \Cref{tab:performance_metric_diff_data} also reveals that $\mathcal{L}_{\text{ROI}}$ is bigger than $\mathcal{L}_{\text{PFD}}$ for most models, which can be expected since the majority of deformation occurs in the ROI, with the exception of the model trained with the robot data. The good performance of such a model in $\mathcal{L}_{\text{ROI}}$ could be attributed to the fact that the contact position is explicitly provided as part of the robot data. The model trained with the combination of images and robot data outperforms those using images or robot data alone, hinting at the effectiveness of our cross-attention mechanism.

\begin{figure}[t]
    \centering
        \includegraphics[width=0.99\linewidth]{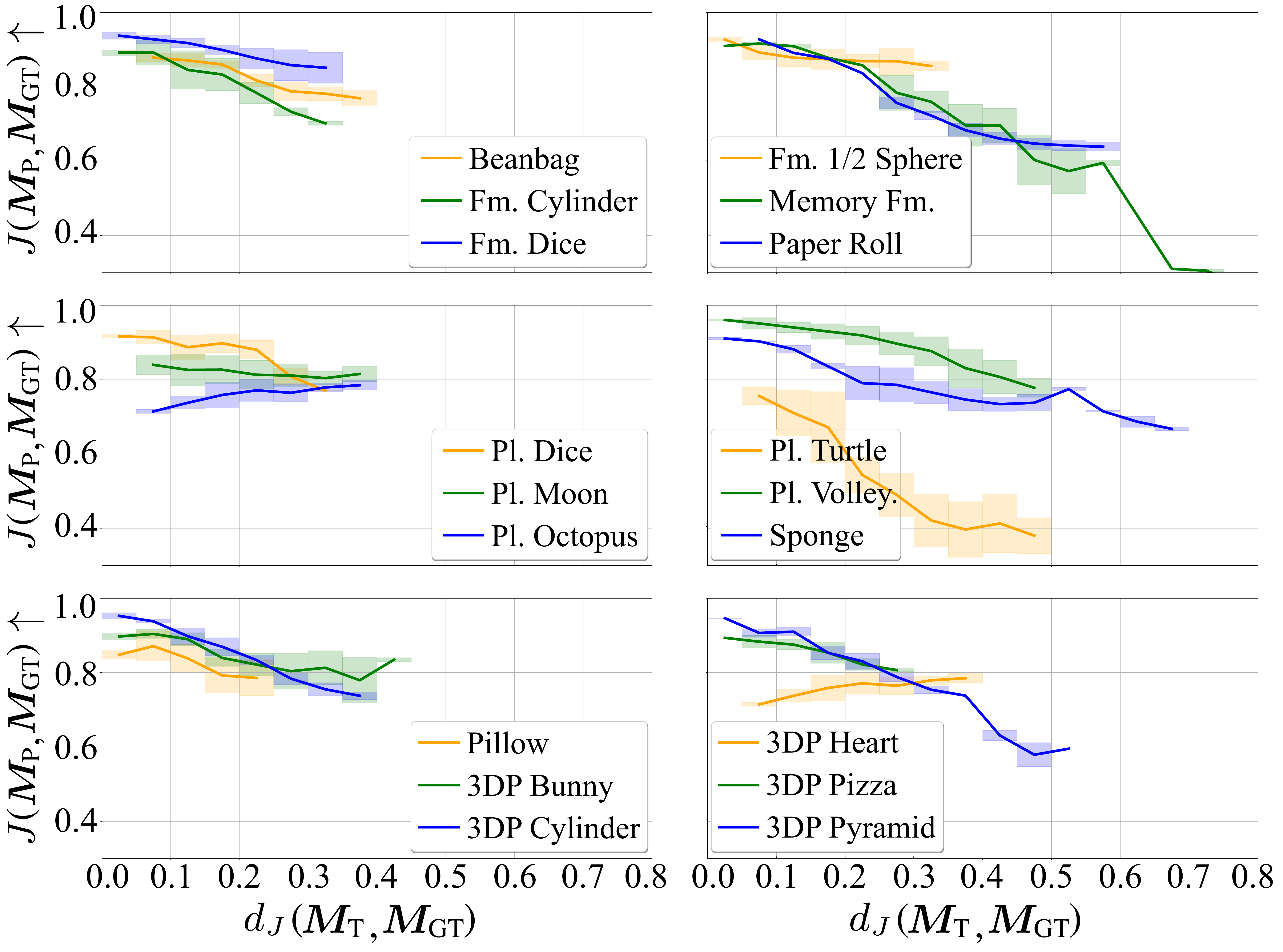}
    %\end{minipage}

    \caption{Validation accuracy for image-based mesh reconstruction, evaluated by Jaccard Index $J$, plotted against the deformation level quantified by Jaccard distance $d_J$. (Fm.:= Foam, Pl.= Plush, 3DP:= 3D Printed, Volley.:= Volleyball )}
    \label{fig:metrics}
\end{figure}

From \Cref{tab:performance_metric_diff_data}, we can observe that the $\text{CD}_{\text{UL1}}$ metric correlates positively with the training losses and it serves as a more intuitive way to evaluate performance in the original scales of the deformable objects.  To analyze the levels of accuracy across multiple objects, we focus on the image-based mesh reconstruction model.
\reb{
\Cref{fig:metrics} illustrates the performance of our model in terms of the Jaccard index $J(\mM_{\text{P}}, \mM_{\text{GT}})$ vs the Jaccard distance, which is defined as $d_{J}(\mM_{\text{T}}, \mM_{\text{GT}}) = 1 - J(\mM_{\text{T}}, \mM_{\text{GT}})$ and indicates the level of deformation of the ground truth mesh with respect to the template mesh. 
A perfect model would predict meshes that result in a horizontal line at $J = 1.0$. For our image-based mesh reconstruction model, we note that the Jarcard Index presents, in general, a slight negative correlation with the Jaccard distance for all objects, indicating that the prediction accuracy decreases for larger deformations. 
The results in \Cref{fig:metrics} also offer insights into the per-object performance, where the best results are obtained for objects like the foam dice and the foam half sphere, while objects like the plush turtle and the memory foam present a more significant challenge for our image-based mesh reconstruction model. 

%The objects where our model performs best are the Foam dice and the Foam Half Sphere. The objects with the worst performance are The plush turtle, the memory foam. 

% Note that the range of values for the Jaccard distance in \Cref{fig:metrics} offers an insight into the different levels of deformation that are present in the dataset.  
}

%\Cref{fig:metrics} shows that the best prediction performance is obtained for the plush moon, having the highest Jaccard Index. Furthermore, we note that, in general, 

\textbf{Reconstruction performance on unseen objects}
\reb{
  To evaluate the generalization of the synthetic point cloud mesh reconstruction model, we use the mean validation performance achieved on the 14 objects that belonged to the training set as a baseline.
  In terms of the point face distance loss $\mathcal{L}_\text{PFD}$, the model demonstrates a good level of generalization for objects like the foam cylinder, the plush dice, and the plush volleyball, compared to the baseline. However, in terms of the unidirectional chamfer distance $\text{CD}_\text{UL1}$, only the plush volleyball outperforms the baseline. As expected, performance tends to degrade on unseen objects.
  Nonetheless, the plush volleyball shows good generalization properties across all metrics, which could be attributed to the fact that it has a similar shape and size to other objects in the training set.
 }

\section{Conclusion}

This paper introduced PokeFlex, a new dataset that captures the behavior of 18 deformable volumetric objects during poking and dropping. The focus is on volumetric objects, while thin clothing items or thin cables are not considered in the dataset. Compared to previously existing datasets, we provide a wider range of paired and annotated data modalities, which are supplemented with data streams from lower-cost camera sensors. 
In an effort to enhance reproducibility, the objects included in our dataset can be either purchased from global providers or 3D printed with our open-source models.
The 3D printed objects also allow for finer control over their expected behavior through knowledge of their material properties and internal structures, especially useful for sim-to-real transfer. 

Using different combinations of the data modalities provided in PokeFlex, \reb{we train and benchmark a list of baseline models, which constitutes the state-of-the-art for the task of multi-object online template-based mesh reconstruction, given the novelty of the multimodal nature of our dataset, to the best of our knowledge}.  Furthermore, we also present a list of suitable criteria for evaluating PokeFlex. 
% We propose a list of evaluation criteria, which the community can use and extend the benchmarks.
% We showed a use case of the dataset training models for online multi-object template-based mesh reconstruction using different combinations of data modalities. 

We are excited about the potential of PokeFlex to inspire new research directions in deformable object manipulation and to serve as a foundational resource for the robotics community. With its rich, multimodal data and its focus on reproducibility, we believe that PokeFlex will drive innovation in both simulation-based and real-world applications of deformable object manipulation. This includes better material parameter identification to fine-tune simulators, viewpoint-agnostic online 3D mesh reconstruction methods, and policy learning for manipulation tasks.  As we continue to expand the dataset and explore new possibilities, we anticipate that PokeFlex will become an invaluable tool for researchers developing next-generation techniques.  We look forward to sharing this dataset with the community and fostering collaborations that push the boundaries of robotics research.

%\vfill

%The preliminary 3D deformation prediction results (\Cref{fig:predictions}) showcase the quality of this pilot dataset. Further improvements in the accuracy of the deformation prediction can potentially be obtained by leveraging other data modalities such as the 3D forces and 3D torques present in the dataset. 

%To encourage the adoption of the PokeFlex dataset by the community, the dataset will be extended to include 3D-printed deformable objects thus enhancing the reproducibility of our results with the release of the corresponding print files. Furthermore, the diversity of deformations applied to the objects will be improved by leveraging additional manipulation strategies such as pinching, dual arm squeezing, lifting, shaking, and tossing. 

%We consider the PokeFlex dataset has the potential to advance the research on deformable objects, enabling a wide range of applications going from online 3D mesh reconstruction, to material parameter identification and policy learning for manipulation tasks. We look forward to making this dataset available for the community. 
\appendix
% \section{Appendix Old}
\subsection{3D Printing Details}
\label{appx:3dprint}
% \HZ{3D print details: material properties, infill pattern and density, number of perimeters, bottom and top layers, print files/models/citations.}
All 3D printed objects were printed using thermoplastic polyurethane (TPU) Filaflex Shore 60A Pro White filament on Prusa MK3S+ and Prusa XL 3D printers equipped with 0.4mm nozzles. The mechanical properties of the filament are presented in \Cref{tab:tpu}.

\begin{table}[ht]
    \caption{Mechanical Properties of Filaflex shore 60A Pro TPU provided by the manufacturer. %\HZ{do we need this table? if yes what info we should provide? @Ronan: I think only the tensile strength is really useful. The stress at 20p is informative}
    }
    \centering
    \footnotesize
    % \begin{adjustbox}{width=\columnwidth}
    \begin{tabular}{lcll}
    \toprule
    % {%
    %     p{4cm}    % Mechanical properties
    %     p{2cm}  % Value
    %     p{3cm}    % Unit
    %     p{5cm}    % Test method according to
    % }
        \textbf{Mechanical properties} & \textbf{Value} & \textbf{Unit} & \textbf{Test method according to}\\ \midrule
        %Hardness &63  &Shore A  &DIN ISO 7619-1 (3s)   \\ 
        Tensile strength &26  &MPa  &DIN 53504-S2  \\ 
        Stress at 20\% elongation &1  &MPa  &DIN 53504-S2  \\ 
        %Tear strength &40  &N/mm  &DIN ISO 34-1Bb  \\ 
        \bottomrule
        
    \end{tabular}
    % \end{adjustbox}
    \label{tab:tpu}
\end{table}

\begin{figure}[h!]
    \centering
    \includegraphics[width=0.6\linewidth]{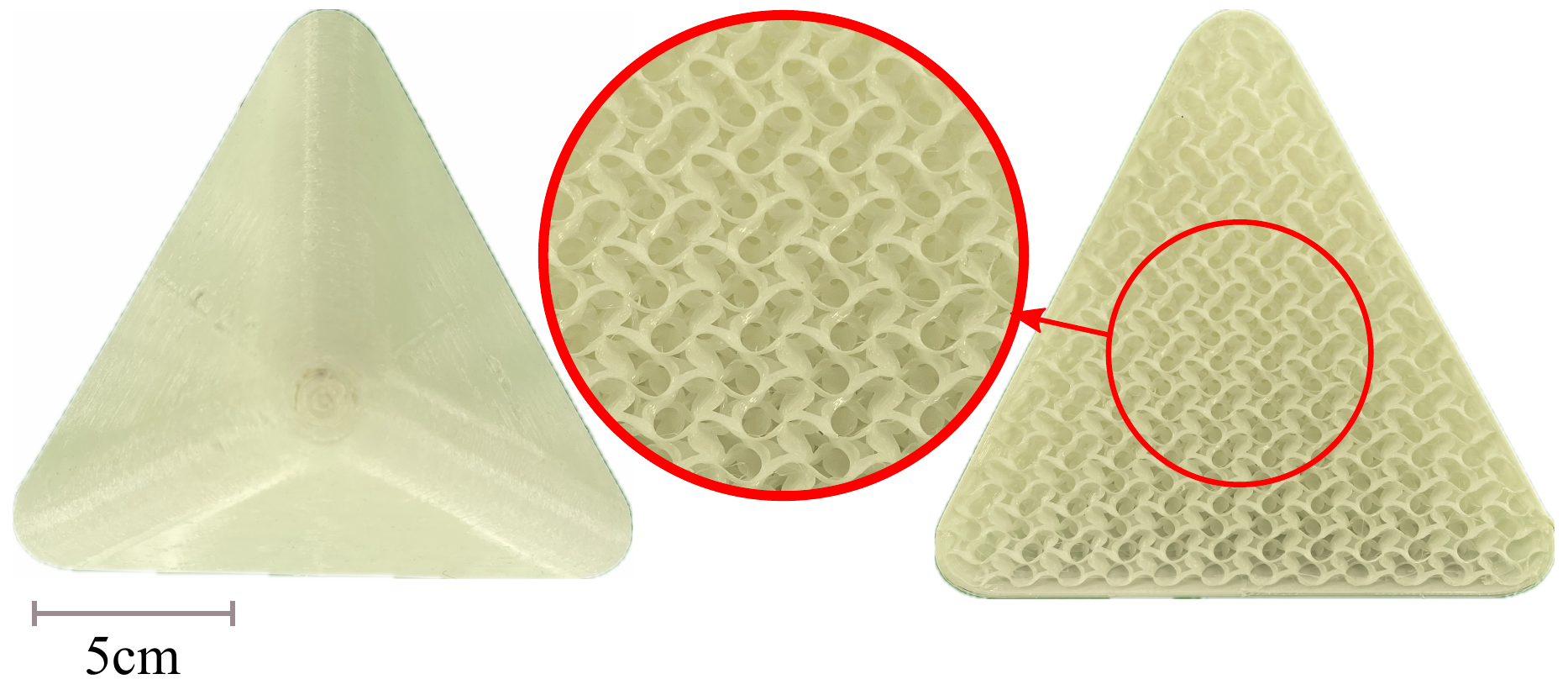}
    \caption{ Top (\textbf{Left}) 
 and bottom (\textbf{Right}) view of 3D printed pyramid, with a close-up view of the gyroid infill pattern.}
    \label{fig:printed_interior}
\end{figure}

The printing parameters of the 3D printed objects are summarized in \Cref{tab:3dprint}, where the infill used for all objects is the isotropic gyroid pattern with uniform properties and behavior in all directions (\Cref{fig:printed_interior}).

\begin{table}[ht]
    \caption{Printing parameters of the 3D printed objects featured in the PokeFlex dataset.}
    \centering
    \footnotesize
    \begin{adjustbox}{width=\columnwidth}
    \begin{tabular}{lccccc}
    \toprule
    % {%
    %     p{3.5cm}    % Object
    %     p{2.5cm}  % Infill Density
    %     p{3cm}    % Layer Thickness
    %     p{1.5cm}    % Perimeters
    %     p{2cm}    % Bottom layers
    %     p{1.5cm}    % Top Layers
    % }
        \textbf{Object} & 
        \begin{tabular}[c]{@{}c@{}}\textbf{Infill density} \\ \textbf{[\%]} \end{tabular}  &
         \begin{tabular}[c]{@{}c@{}}\textbf{Layer thick-} \\ \textbf{ness [mm]} \end{tabular}  &
        
        % \textbf{Perimeters} & 

        \begin{tabular}[c]{@{}c@{}}\textbf{Peri-} \\ \textbf{meters} \end{tabular}  &
        
        \begin{tabular}[c]{@{}c@{}}\textbf{Bottom} \\ \textbf{layers} \end{tabular}  &
        \begin{tabular}[c]{@{}c@{}}\textbf{Top} \\ \textbf{layers} \end{tabular} 
        \\ \midrule
        Bunny \cite{Turk1994bunny} &10  &0.2  &3  &3  &3  \\ 
        Cylinder &10  &0.15  &2  &3  &3  \\ 
        Heart \cite{noor20193dheart} &10  &0.2  &3  &3  &3  \\ 
        Pizza &10  &0.2  &3  &3  &3 \\ 
        Pyramid  &8  &0.2  &3  &3  &3  \\ \bottomrule
    \end{tabular}
    \end{adjustbox}
    \label{tab:3dprint}
\end{table}

\subsection{Properties of Featured Objects}
\label{appx:object}
% \HZ{are we gonna provide purchase links either in appendix or say we provide the links on our website..?}

In \Cref{tab:objects}, we summarize the physical properties and the number of frames per object. The Frames column of the table presents the total captured frames of the poking sequences for each object, and the Deformations column gives the number of active poking frames after the data curation, i.e., discarding the frames where the robot arm is not in contact with the objects. It is worth noting that we report only the effective number of paired time frames in our table, in contrast to the total number of samples, which is computed as the number of time frames multiplied by the number of cameras. 

\begin{table}[h]
    \caption{\reb{Physical properties of objects featured in the PokeFlex dataset. Dimensions of sphere-like objects are described by their diameter (D). Cylinder-like objects are characterized by their diameter (D) and height (H). For objects with irregular or complex shapes, dimensions are provided using a bounding box defined by length (L), width (W), and height (H). Stiffness of the objects is estimated according to the method described in \Cref{sec:results_dataset}.}}
    \centering
    \footnotesize
    \begin{adjustbox}{width=\columnwidth}
    \begin{tabular}{%
        % p{3cm}    % Object
        % p{2cm}    % Weight
        % p{3cm}    % Dimensions
        % p{3cm}    % Est. stiffness
        % p{1.5cm}    % Frames
        % p{2cm}    % Deformations
        llllll
    }
    \toprule
        \textbf{Object} & 
        \begin{tabular}[c]{@{}l@{}}\textbf{Weight} \\ \textbf{[g]} \end{tabular}  &
        \textbf{Dimensions [cm]} & 
        
        \begin{tabular}[c]{@{}l@{}}\textbf{Est. stiffness} \\ \textbf{[N/m]} \end{tabular}  &
        
        \begin{tabular}[c]{@{}l@{}}\textbf{Total} \\ \textbf{frames} \end{tabular}  &

        \begin{tabular}[c]{@{}l@{}}\textbf{Frames} \\ \textbf{in contact} \end{tabular}
        
        \\ \midrule
        Beanbag & 184 & DxH: 26x9 & 523 & 1084 & 363\\ 
        Foam cylinder & 153 & DxH: 12x29 & 250 & 990 & 407 \\ 
        Foam dice & 140 & L: 15.5 & 748 & 1220 & 738 \\ 
        Foam half sphere & 41 & D: 15 & 1252 & 939 & 546\\ 
        Memory foam & 213 & LxWxH: 17.5x8.5x7 & 395 & 958 & 282 \\ 
        Pillow & 975 & LxWxH: 58x50x10 & 474 & 1085 & 565\\ 
        Plush dice & 340 & L: 22 & 149 & 1259 & 567 \\ 
        Plush moon & 151 & D: 17 & 366 & 959 & 517\\ 
        Plush octopus & 130 & LxWxH: 22x22x11 & 325 & 1085 & 525 \\ 
        Plush turtle & 194 & LxWxH: 35x30x10 & 1035 & 930 & 427\\ 
        Plush volleyball & 303 & D: 22 & 323 & 939 & 488\\ 
        Sponge & 28 & LxWxH: 22x12x6.1 & 1045 & 775 & 490 \\ 
        Toilet paper roll & 134 & DxH: 10.5x9.5 & 2156 & 920 & 411 \\ 
        3D printed bunny & 105 & LxWxH:13x9x15 & 950 & 1117 & 520 \\ 
        3D printed cylinder & 223 & DxH: 10x20 & 585 & 1020 & 574 \\ 
        3D printed heart & 100 & LxWxH: 16x9x10 & 1198 & 940 & 444 \\ 
        3D printed pizza & 68 & LxWxH:18x15x3 & 884 & 958 & 360\\ 
        3D printed pyramid & 48 & LxWxH: 14.5x14.5x7 & 861 & 899 & 386 \\ \bottomrule
    \end{tabular}
    \end{adjustbox}
    \label{tab:objects}
\end{table}

For the dropping protocol, we recorded 3 sequences of 1 second at 60 fps for each object, summing up to 180 time frames per object. %\Cref{fig:drop} shows two additional reconstructed deformed mesh sequences for dropping the foam cylinder and the pillow, respectively.

% \begin{figure}[h!]
% \centering
% \vspace{0.3cm}
% \includegraphics[width=0.99\linewidth]{images/drop_v4.pdf}
% % \vspace{0.3cm}
% \caption{Sample of mesh reconstructions of foam cylinder (\textbf{Top}) and pillow (\textbf{Bottom}) for a dropping sequence.}
% \label{fig:drop}
% \end{figure}

\subsection{Training Details}
\label{appx:train_detail}
The hyperparameters used to train the models in \Cref{sec:results_reconstruction} are listed in \Cref{tab:training_details}.

\begin{table}[h]
    \centering
    \caption{Training hyperparameters.}
    \footnotesize % or use \footnotesize for even smaller text
    %\begin{adjustbox}{width=\columnwidth}
    %\begin{tabularx}{\textwidth}{X X}
    \begin{tabular}{ll}
        \toprule
        \textbf{Hyperparameters} & \textbf{Value} \\ \midrule
        Learning rate & 1e-4 \\ 
        Batch size & 32 \\ 
        Optimizer & Adam \\ 
        Weight decay & 0.0 \\ 
        Learning rate scheduler & Cosine\\ 
        % Maximum number of iterations & 10000\\ \hline
        Minimum learning rate & 1e-7\\ 
        Epochs & 200 \\ \bottomrule
    \end{tabular}
    %\end{tabularx}
    %\end{adjustbox}
    \label{tab:training_details}
\end{table}

\subsection{Inference Speed for Different Input Data Modalities}
\label{appx:inference}

\Cref{tab:inference_speed} shows the measured inference rates for six models that combine different input data modalities. The rate is tested with an AMD Ryzen 7900 x 12 Core Processor CPU and NVIDIA GeForce RTX 4090 GPU with 24GB memory. 
\begin{table}[h!]
    \centering
    \caption{Inference rates for the models that ingest different input modalities }
    \footnotesize % or use \footnotesize for even smaller text
    %\begin{adjustbox}{width=\columnwidth}
    %\begin{tabularx}{\textwidth}{X X}
    \begin{tabular}{lc}
        \toprule
        \textbf{Input} & \textbf{Inference Rate} \\ \midrule
        Images & 115 Hz \\ 
        \reb{Robot data} & 185 Hz \\ 
        Images + robot data & 110 Hz \\ 
        Point clouds (5000 points) & 33 Hz \\ 
        Point clouds (5000 points) + robot data & 33 Hz \\ 
        Point clouds (100 points) & 180 Hz \\
        \bottomrule
    \end{tabular}
    %\end{tabularx}
    %\end{adjustbox}
    \label{tab:inference_speed}
\end{table}

%%%%%%%%%%%%%%%%%%%%%%%%%%%%%%%%%%%%%%%%%%%%%%%%%%%%%%%%%%%%%%%%%%%%%%%%%%%%%%%%

%%%%%%%%%%%%%%%%%%%%%%%%%%%%%%%%%%%%%%%%%%%%%%%%%%%%%%%%%%%%%%%%%%%%%%%%%%%%%%%%

%%%%%%%%%%%%%%%%%%%%%%%%%%%%%%%%%%%%%%%%%%%%%%%%%%%%%%%%%%%%%%%%%%%%%%%%%%%%%%%%
%\section*{APPENDIX}

%\clearpage
%\input{chapters/90_appendix}

% \section*{ACKNOWLEDGMENT}
% This article is supported by the SDSC Grant entitled 'C22-08: Data-Driven Inference of Mesh-based Representations for Deformable Objects from Unstructured Point Clouds'

%%%%%%%%%%%%%%%%%%%%%%%%%%%%%%%%%%%%%%%%%%%%%%%%%%%%%%%%%%%%%%%%%%%%%%%%%%%%%%%%

%\addtolength{\textheight}{-17cm}   % This command serves to balance the column lengths

\bibliographystyle{IEEEtran}
\bibliography{root.bib}

%\clearpage
%\input{chapters/90_appendix}
%%%%%%%%%%%%%%%%%%%%%%%%%%%%%%%%%%%%%%%%%%%%%%%%%%%%%%%%%%%%%%%%%%%%%%%%%%%%%%%%
\end{document}